\documentclass[format=manuscript, nonacm]{acmart}
\usepackage{caption}
\usepackage{verbatim}
\usepackage{tikz}
\usepackage{xcolor}
\usepackage{bbding} 
\usepackage{microtype}
\usepackage{colortbl}
\usepackage{subcaption}
\usepackage{booktabs} 

\usepackage{titlesec}

\titleformat{\subsubsection} 
  {\normalfont\normalsize\bfseries} 
  {\thesubsection.\arabic{subsubsection}} 
  {1em} 
  {} 

\usepackage{microtype}
\usepackage{graphicx}
\usepackage{multirow}
\usepackage{comment}
\usepackage{booktabs} 
\usepackage{tikz}
\usetikzlibrary{positioning, arrows.meta, calc}

\renewcommand\footnotetextcopyrightpermission[1]{} 
\AtBeginDocument{%
  }
    
\setcopyright{acmlicensed}
\copyrightyear{2018}
\acmYear{2018}
\acmDOI{XXXXXXX.XXXXXXX}
\acmConference[Conference acronym 'XX]{Make sure to enter the correct
  conference title from your rights confirmation email}{June 03--05,
  2018}{Woodstock, NY}

\acmISBN{978-1-4503-XXXX-X/2018/06}

\author{Fangming Cui}
\affiliation{%
  \institution{Beijing}
  \city{Beijing}
  \country{China}
  }

\author{Ruixiao Zhu}
\affiliation{%
  \institution{Shanghai}
  \city{Beijing}
  \country{China}
  }

\author{Cheng Fang}
\affiliation{%
  \institution{Shanghai}
  \city{Beijing}
  \country{China}
  }

\author{Sunan Li}
\affiliation{%
  \institution{Shanghai}
  \city{Beijing}
  \country{China}
  }

\author{Jiahong Li}
\affiliation{%
  \institution{Shanghai}
  \city{Beijing}
  \country{China}
  }

\begin{document}

\title{Rethinking Agentic Reinforcement Learning In Large Language Models}



\begin{abstract}
Reinforcement Learning (RL) has traditionally focused on training specialized agents to optimize predefined reward functions within narrowly defined environments. However, the advent of powerful Large Language Models (LLMs) and increasingly complex, open-ended tasks has catalyzed a paradigm shift towards agentic paradigms within RL. This emerging framework extends beyond traditional RL by emphasizing the development of autonomous agents capable of goal-setting, long-term planning, dynamic strategy adaptation, and interactive reasoning in uncertain, real-world environments. Unlike conventional approaches that rely heavily on static objectives and episodic interactions, LLM-based Agentic RL incorporates cognitive-like capabilities such as meta-reasoning, self-reflection, and multi-step decision-making directly into the learning loop. In this paper, we provide a deep insight for looking the conceptual foundations, methodological innovations, and effective designs underlying this trend. Furthermore, we identify critical challenges and outline promising future directions for building LLM-based Agentic RL. 
\end{abstract}

\ccsdesc[500]{Computing methodologies~Artificial intelligence}
\ccsdesc[500]{Computing methodologies~Machine learning}
\keywords{Agentic RL, LLMs, Designs}


\maketitle

\makeatletter
\definecolor{forcePurple}{RGB}{128,128,128} 

\let\oldcite\cite

\renewcommand{\cite}[1]{%
  \textcolor{forcePurple}{\oldcite{#1}}%
}

\ifdefined\citep
  \let\oldcitep\citep
  \renewcommand{\citep}[1]{\textcolor{forcePurple}{\oldcitep{#1}}}
\fi
\ifdefined\citet
  \let\oldcitet\citet
  \renewcommand{\citet}[1]{\textcolor{forcePurple}{\oldcitet{#1}}}
\fi

\makeatother

\section{Introduction}
Marking a pivotal shift from static text generation to dynamic decision-making, the advent of Agentic Reinforcement Learning (RL) fundamentally redefines the operational paradigm of Large Language Models (LLMs)~\cite{hu2026seeupo}. While conventional LLMs function primarily as sophisticated auto-regressive predictors, excelling at next-token prediction within the confines of static prompts, the integration of RL principles transforms these models from mere passive responders into proactive agents. Within this novel framework, LLMs are formulated not just as text generators, but as policies $\pi_\theta(a|s)$ operating within Partially Observable Markov Decision Processes~\cite{curtis2025Llm-guided}. This formulation is crucial because it acknowledges the inherent uncertainty and sequential dependency of real-world tasks. Unlike conventional supervised fine-tuning (SFT) or even standard Reinforcement Learning from Human Feedback (RLHF), which often treat each response as an independent instance hinging on single-step preferences~\cite{ouyang2022training}, Agentic RL situates the LLM within a continuous feedback loop. By shifting the optimization objective from instantaneous rewards to the cumulative return of entire trajectories~\cite{Agent-Omit}, this approach enables the model to engage in multi-step interactions where the consequences of early actions propagate through time. 
Such a paradigm equips agents with three critical cognitive faculties: long-horizon planning, strategic tool utilization, and persistent memory. Long-horizon planning allows the agent to decompose a monolithic user query into a sequence of manageable sub-goals, executing them in order without losing sight of the ultimate objective. Tool utilization involves the agent learning when and how to interface with external APIs, calculators, or search engines, thereby overcoming the knowledge cut-off and factual hallucination issues inherent to standalone LLMs~\cite{choi2025reactree}. Meanwhile, persistent memory mechanisms, ranging from vector databases to in-context summarization, enable the agent to retain context across multiple turns of interaction~\cite{nakano2021webgpt}, facilitating a form of experiential learning that mirrors biological cognition~\cite{zhong2024memorybank}. Through this synergistic architecture, Agentic RL effectively bridges the chasm between passive language models and truly autonomous, self-improving systems capable of meta-reasoning.
Unleashed by these capabilities, a new frontier of complex application scenarios emerges, moving far beyond the scope of simple chatbot interactions~\cite{zhu2025convsearch}. In the realm of software engineering~\cite{badertdinov2025swe,golubev2025training,takerngsaksiri2025human,yan2025re,zeng2025satori} , these agents evolve from merely generating simple code snippets~\cite{le2022coderl} to autonomously managing entire software repositories~\cite{dai2024process,chen2025r1,dou2024stepcoder,gehring2025rlef}; they can now perform complex debugging cycles, write unit tests, code~\cite{anthropic2025claude,chen2025breaking}, and even optimize algorithms based on runtime performance metrics, effectively acting as autonomous software developers. Similarly, in scientific discovery~\cite{yamada2025ai}, Agentic RL powers systems that can automate the research lifecycle: synthesizing insights from thousands of papers, designing computational experiments, and analyzing multivariate datasets. In the domain of web navigation~\cite{geng2025webwatcher,gou2025mind2web,wei2025webagent}, these agents transcend basic information retrieval by performing sophisticated tasks such as online shopping (comparing prices and checking reviews)~\cite{koh2024visualwebarena,li2025webthinker,vattikonda2025how}, booking travel itineraries, or managing digital administrative tasks that require filling out forms and navigating multi-page websites~\cite{hao2025rl}. Furthermore, the applicability of LLM-based Agentic RL extends into highly specialized verticals, from mathematical reasoning~\cite{asperti2025thinking,mai2025agent,ren2025deepseek,shao2022deepseek} where agents tackle competition-level problems by interleaving natural language reasoning with symbolic computation, to financial analysis~\cite{yu2024fincon}, where they serve as portfolio management assistants that continuously monitor market data. Perhaps most compelling is the emergence of embodied AI~\cite{gao2025octonav,guo2024luban,kang2025viki,shridhar2021alfworld}, where the agent acts as the cognitive core for physical robots or virtual avatars~\cite{song2025maniplvm,wang2024voyager,zala2024envgen,zhao2025embodied}. Here, high-level linguistic commands (e.g., “fetch the coffee mug from the kitchen”) are translated into sequences of low-level motor controls or navigation plans in simulated or real-world environments. GUI~\cite{kumar2025trishul,lian2025ui,liu2025infi,liu2025infigui,lu2025ui,luo2025gui,qin2025ui,shi2025mobilegui,yan2023gpt4v,zhang2025tongui,zheng2025naturegaia} is also a very important task scenario in human society.
While prior surveys have provided comprehensive overviews of LLMs, Agents, and Reinforcement Learning, either individually or in pairwise combinations, few have integrated all three domains effectively. Existing multi-domain studies include work~\cite{lin2025comprehensivesurvey}, which targets search and recommendation systems, and work2~\cite{zhang2025landscape}, which offers a panoramic, library-centric survey with a breadth-first approach. 

\textbf{Our Contributions.} In this paper, we provide a deep looking for formulas, effective designs and innovative skills, adopting a concise format to rapidly assimilate the current state of the industry.

\section{Designs} 

\subsection{Paradigm of LLM-Based Agents}
We specifically introduce the formulas for each component of agent, followed closely by highlighting the contributions of key works. In constructing autonomous agents capable of complex task execution, it is imperative to adopt a systematic framework that transcends the capabilities of standalone Large Language Models (LLMs). An intelligent agent operates within a control-theoretic loop comprising core components: {Action}, {Planning}, {Memory}, and {Tools}. 

\textbf{Action (The Interface of Intervention).}
Actions are the sole means by which the agent influences the environment. 
Action Space: $\mathcal{A} = \{a^{(1)}, a^{(2)}, \dots, a^{(N)}\}$, common in game-playing agents.
Continuous Action Space: $\mathcal{A} \subseteq \mathbb{R}^d$, typical in robotics, where $d$ is the dimensionality of the action vector.
The agent's behavior is dictated by its \textit{policy} $\pi$, a mapping from states to actions. To evaluate action quality, the Q-function is defined.
\begin{equation}
    \begin{aligned}
        a_t \sim \pi_\theta(a | s_t),
    \end{aligned}
\end{equation}
\begin{equation}
    \begin{aligned}
\pi_\theta(a | s) = \mathcal{N}(\mu_\theta(s), \Sigma_\theta(s)),
    \end{aligned}
    \end{equation}
\begin{equation}
    \begin{aligned}
a_t = \mu_\theta(s_t),
    \end{aligned}
    \end{equation}
\begin{equation}
    \begin{aligned}
Q^\pi(s, a) = \mathbb{E}_\pi \left[ \sum_{k=0}^\infty \gamma^k R_{t+k} \,\middle|\, S_t = s, A_t = a \right],
\end{aligned}
    \end{equation}
\begin{equation}
    \begin{aligned}
Q^*(s, a) = \mathbb{E}_{s'} \left[ r + \gamma \max_{a'} Q^*(s', a') \right].
\end{aligned}
    \end{equation}
Predefined action spaces, while effective in narrow domains, fundamentally constrain LLM agents in open-ended scenarios by limiting planning capabilities and demanding impractical human effort for exhaustive enumeration. Addressing this, a dynamic framework~\cite{Dynasaur} is proposed, empowering agents to generate and execute programs ad hoc, with actions accumulating for future reuse. However, standard tuning methods, whether via supervised fine-tuning on ReAct trajectories or preference optimization, suffer from limited exploration, causing over-commitment to suboptimal actions. To counter this, the SAND framework~\cite{SAND} is introduced, enabling explicit deliberation over candidates via self-consistency sampling and execution-guided critique. Complementing this, planning hallucinations arising from a lack of built-in action knowledge are targeted by KnowAgent~\cite{zhu2025knowagent}, which constrains planning trajectories using an action knowledge base. Finally, the derivation of actionable principles from data is tackled by the PRAct framework~\cite{Pract}, whose core innovation lies in using textual gradients from a reflective engine. 

\begin{figure*}[t]
\centerline{\includegraphics[width=0.55\columnwidth]{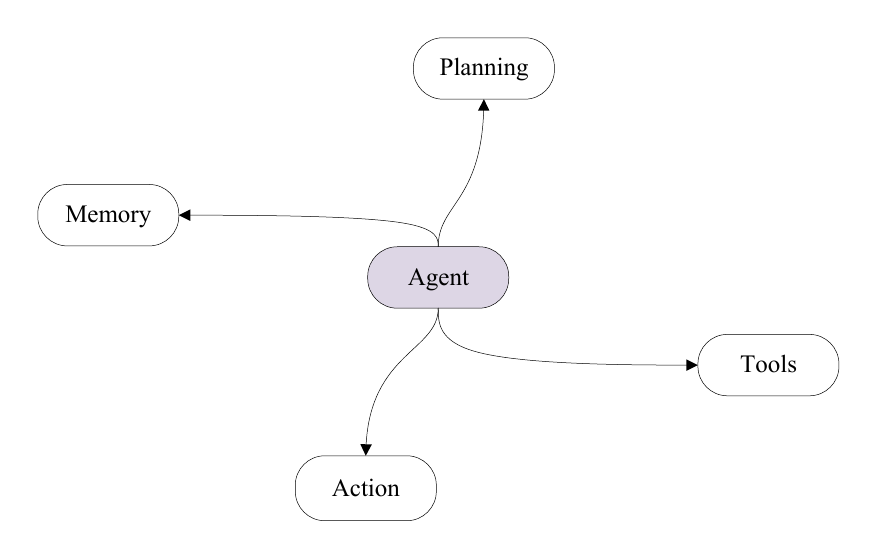}}
\caption{Agent.}
\label{agent}
\end{figure*} %

\textbf{Planning (Prospective Reasoning).} Planning involves using an internal model to simulate future trajectories without real-world execution.
MCTS builds a search tree via simulations. The selection phase utilizes the Upper Confidence Bound (UCB1) algorithm,  where $Q(s,a)$ is the mean value, $N(\cdot)$ are visit counts, and $c$ is an exploration constant.
The agent learns a dynamics model $\hat{f}_\theta$ and it then solves a finite-horizon optimization problem.
\begin{equation}
    \begin{aligned}
a_t = \arg\max_a \left( Q(s, a) + c \sqrt{\frac{\ln N(s)}{N(s, a) + \epsilon}} \right),
\end{aligned}
    \end{equation}
\begin{equation}
    \begin{aligned}
\hat{s}_{t+1} = \hat{f}_\theta(s_t, a_t),
\end{aligned}
    \end{equation}
\begin{equation}
    \begin{aligned}
\min_{a_t, \dots, a_{t+H}} \sum_{k=0}^{H} \gamma^k \hat{r}_{t+k} \quad \text{s.t.} \quad \hat{s}_{t+k+1} = \hat{f}_\theta(\hat{s}_{t+k}, a_{t+k}).
\end{aligned}
    \end{equation}
Despite exhibiting impressive performance across a vast array of tasks, Large Language Models (LLMs) frequently encounter difficulties when tasked with multi-step reasoning or goal-directed planning. Drawing inspiration from the human brain, where planning emerges from component processes tied to specific neural substrates, such as conflict monitoring, state prediction, and task coordination, researchers identify a critical deficit in LLMs: while individually capable of these functions, autonomous coordination toward a goal remains a challenge. This limitation motivates the proposal of the work~\cite{webb2025brain}. Similarly, while paradigms like ReAct instruct LLMs to plan explicitly before each action, the computational infeasibility of perpetual planning and the performance degradation it causes on long-horizon tasks are demonstrated by researchers, who conversely show that abstaining from planning imposes its own ceiling on performance. Addressing this dichotomy, a framework~\cite{paglieri2025learning} formalizing dynamic planning is introduced, empowering LLM agents to flexibly allocate test-time compute. Further tackling the generalization gap induced by supervised fine-tuning, which often reduces models to memorizing trajectories, researchers present work~\cite{lu2025pilotrl}, an adaptive global plan-based paradigm that synergizes high-level guidance with execution to facilitate long-horizon decision-making.

\textbf{Memory (Temporal Context Integration).}
Memory allows the agent to overcome partial observability by storing and retrieving historical information.
Recurrent networks maintain a hidden state $h_t$, the Long Short-Term Memory (LSTM) cell updates are defined as:
\begin{align*}
i_t &= \sigma(W_i \cdot [h_{t-1}, x_t] + b_i) & \text{(Input Gate)} \\
f_t &= \sigma(W_f \cdot [h_{t-1}, x_t] + b_f) & \text{(Forget Gate)} \\
\tilde{C}_t &= \tanh(W_C \cdot [h_{t-1}, x_t] + b_C) & \text{(Candidate)} \\
C_t &= f_t \odot C_{t-1} + i_t \odot \tilde{C}_t & \text{(Cell State)} \\
o_t &= \sigma(W_o \cdot [h_{t-1}, x_t] + b_o) & \text{(Output Gate)} \\
h_t &= o_t \odot \tanh(C_t) & \text{(Hidden State)}
\end{align*}
Deep Q-Networks (DQN) utilize a replay buffer $\mathcal{D}$ and training samples mini-batches $\mathcal{B}$ uniformly. Long-term knowledge is stored in a vector database. 
\begin{equation}
    \begin{aligned}
\mathcal{D} = \{(s_i, a_i, r_i, s'_i, d_i)\}_{i=1}^N,
\end{aligned}
    \end{equation}
\begin{equation}
    \begin{aligned}
\mathcal{B} \sim \mathcal{U}(\mathcal{D}).
\end{aligned}
    \end{equation}
Despite the impact of Large Language Models (LLMs) on AI interactions and their broad task proficiency, a critical constraint persists: the absence of a native long-term memory mechanism. This deficiency proves particularly detrimental in sustained interaction scenarios, such as personal companionship or secretarial assistance, motivating the proposal of work~\cite{zhong2024memorybank}. While recent attempts to augment LLMs with external memory banks address the statelessness imposed by finite context windows, they typically rely on static, heuristic-driven pipelines lacking learned control over storage and retrieval. Addressing this, a novel work~\cite{yan2025memory} is presented as an RL framework that empowers LLMs to actively manage external memory via specialized agents. Complementing this, the inefficiency of full-context prompting, which appends all historical turns irrespective of relevance, causing unbounded memory growth and performance degradation, is tackled by work~\cite{zhou2025mem1}, an end-to-end RL framework enabling agents to operate with constant memory across extended multi-turn tasks. Addressing the limitations of static memory, a scalable, memory-centric architecture named Mem0~\cite{chhikara2025mem0} is introduced, designed to dynamically extract and consolidate salient information from ongoing dialogues. Building upon this foundation, an enhanced variant is further proposed, one that leverages graph-based representations to explicitly capture the complex relational structures inherent in conversation.

\textbf{Tools (External Capability Extension).}
Tools allow the agent to transcend parametric knowledge limitations by interfacing with external APIs or devices.
A tool $T_i$ is treated as an external black-box function, where $x_t$ is the structured input (e.g., JSON arguments) generated by the agent, and $y_t$ becomes the subsequent observation $o_{t+1}$.
Selecting the appropriate tool can be framed as a classification problem,  $\phi(s_t)$ encodes the current context, the Reasoning + Acting (ReAct) paradigm interleaves thoughts and actions.
\begin{equation}
    \begin{aligned}
y_t = T_i(x_t; \theta_T),
\end{aligned}
    \end{equation}
\begin{equation}
    \begin{aligned}
p(T_i | s_t) = \text{Softmax}(W_T \cdot \phi(s_t)),
\end{aligned}
    \end{equation}
\begin{equation}
    \begin{aligned}
\text{Thought}_t \rightarrow \text{Action}_t (\text{Tool Call}) \rightarrow \text{Observation}_{t+1} \rightarrow \text{Thought}_{t+1}.
\end{aligned}
    \end{equation}
An interleaved generation of reasoning traces and task-specific actions is explored by researchers, a paradigm that fosters a powerful synergy: while reasoning facilitates the induction and updating of action plans, actions themselves interface with external environments to gather critical information. Applied to a diverse set of language and decision-making tasks, this approach~\cite{yao2023react}, demonstrates clear superiority over state-of-the-art baselines, offering enhanced human interpretability compared to non-interactive methods. Extending this line of inquiry, work~\cite{chen2023fireact} is proposed to fine-tune LMs using trajectories from multiple tasks and prompting strategies, revealing that data diversity is key to improving agent robustness and generalization. These findings collectively establish the comprehensive benefits of agent fine-tuning, moving beyond mere scaling effects to address efficiency and cost. 
Addressing the limitations of pure RL-trained reasoning models (e.g., DeepSeek R1) in structured problem-solving, where computational tools hold distinct advantages, researchers introduce ReTool~\cite{feng2025retool}. This framework bridges the gap by integrating tool usage into long-form reasoning, featuring a dynamic interleaving of real-time code execution with natural language, alongside an automated RL paradigm that teaches the model when and how to invoke tools based on outcome feedback.

\subsection{Paradigm of Reinforcement Learning} 
We provide a deep looking to analyze and observe the following works. First, we introduce its formulas and principles of key works, and finally elaborate based on its derivative style.

\textbf{Proximal Policy Optimization (PPO).}
PPO~\cite{schulman2017proximal}  is a widely used policy gradient algorithm in reinforcement learning, known for its stability and ease of implementation. Its key feature is the clipped objective function, which limits policy updates to prevent large, destabilizing changes. This "proximal" update mechanism ensures that the new policy stays close to the old one, improving training reliability. PPO typically uses an actor-critic architecture, where the actor updates the policy and the critic estimates value functions. It is sample-efficient and performs well across various continuous and discrete control tasks. Variants like PPO-Clip and PPO-Penalty offer different ways to enforce policy constraints. Overall, PPO balances performance and simplicity, making it a popular choice for both research and practical applications in robotics, gaming, and LLM fine-tuning.
The formula represents the Proximal Policy Optimization (PPO) objective $\mathcal{J}_{\text{PPO}}(\theta)$, which is defined as the expectation over states $x$ from dataset $\mathcal{D}$ and actions $y$ sampled from the old policy $\pi_{\theta_{\text{old}}}$, averaging over the timesteps $|y|$; specifically, it computes the minimum between the product of the probability ratio $w_t(\theta)$ and the clipped ratio, both multiplied by the advantage estimate $\hat{A}_t$, thereby constraining policy updates within a trust region controlled by the hyperparameter $\varepsilon$ to stabilize training. The PPO loss is defined as:
\begin{equation}
    \begin{aligned}
        \mathcal{J}_{\mathrm{PPO}}(\theta)=\mathbb{E}_{x \sim \mathcal{D}, y \sim \pi_{\theta_{\mathrm{old}}}(\cdot \mid x)}\left[\frac{1}{|y|} \sum_{t=1}^{|y|} \min \left(w_{t}(\theta) \widehat{A}_{t}, \operatorname{clip}\left(w_{t}(\theta), 1-\varepsilon, 1+\varepsilon\right) \widehat{A}_{t}\right)\right],
    \end{aligned}
\end{equation}
where
\begin{equation}
    \begin{aligned}
        w_{t}(\theta)=\frac{\pi_{\theta}\left(y_{t} \mid x, y_{<t}\right)}{\pi_{\theta_{\text {old }}}\left(y_{t} \mid x, y_{<t}\right)}.
    \end{aligned}
\end{equation}

\textbf{Direct Preference Optimization (DPO).}
DPO~\cite{rafailov2023direct}  is a method for fine-tuning large language models (LLMs) to align with human preferences. Unlike traditional RLHF (Reinforcement Learning from Human Feedback) methods that require training a separate reward model and using complex reinforcement learning, DPO directly optimizes the model by leveraging a simple binary cross-entropy loss on preference data. This approach eliminates the need for a reward model and complex RL loops, making the training process more stable, efficient, and computationally less expensive. DPO has become a popular choice for aligning LLMs due to its simplicity and effectiveness in improving model outputs based on human feedback.
Let $D = \{(y_w, y_l)\}$ represent a dataset of pairwise preferences, where $y_w$ denotes the preferred (winning) response and $y_l$ denotes the dispreferred (losing) response. $\pi_{\text{ref}}$ is the reference policy (typically the initial SFT model) and $\beta$ is a hyperparameter controlling the regularization strength. 
\begin{equation}
    \begin{aligned}
\mathcal{J}_{\mathrm{DPO}}(\theta)=\mathbb{E}_{\left(x, y_{w}, y_{l}\right) \sim \mathcal{D}}\left[-\log \sigma\left(\beta\left(\log s_{w}(\theta)-\log s_{l}(\theta)\right)\right)\right],
    \end{aligned}
\end{equation}
where
\begin{equation}
    \begin{aligned}
s_{w}(\theta)=\frac{\pi_{\theta}\left(y_{w} \mid x\right)}{\pi_{\mathrm{ref}}\left(y_{w} \mid x\right)}, s_{l}(\theta)=\frac{\pi_{\vartheta}(y_{l} \mid x)}{\pi_{\mathrm{ref}}(y_{l} \mid x)}
    \end{aligned}.
\end{equation}

\begin{figure*}[t]
\centerline{\includegraphics[width=1.03\columnwidth]{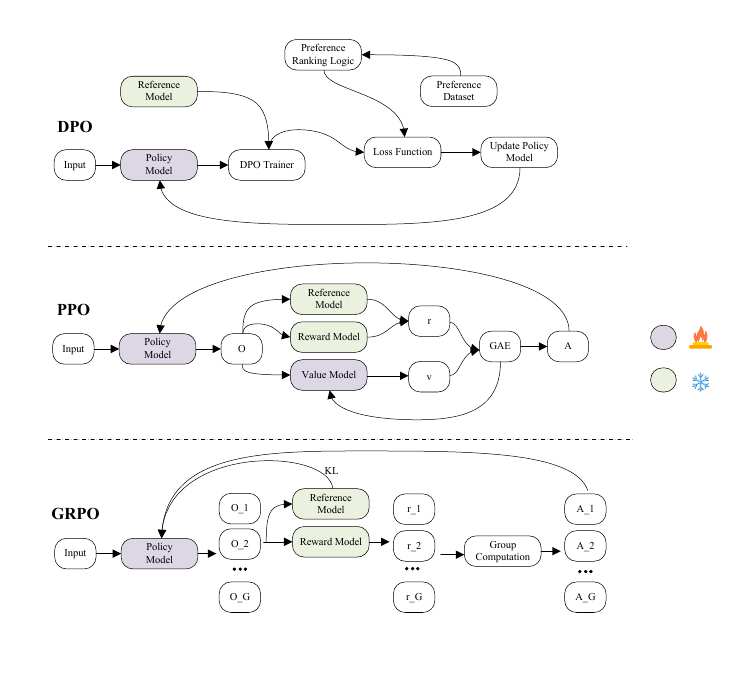}}
\caption{Demonstration of DPO, PPO and GRPO.}
\label{rader}
\end{figure*} 

\textbf{Group Relative Policy Optimization (GRPO).}
GRPO~\cite{shao2022deepseek} is an advanced reinforcement learning algorithm proposed by DeepSeek to enhance the reasoning capabilities of Large Language Models (LLMs). It serves as an efficient alternative to Proximal Policy Optimization (PPO). The core innovation of GRPO is the elimination of the resource-intensive Critic (value) model. Instead of relying on absolute value estimation, GRPO samples a group of outputs for each input prompt and calculates the relative advantage by normalizing rewards within the group. This "group relative" mechanism significantly reduces memory usage and computational overhead while maintaining training stability through clipping mechanisms and KL divergence constraints. GRPO is particularly effective for complex reasoning tasks, such as mathematical problem-solving and code generation. For each query $x$, GRPO generates a total of $G$ responses. Subsequently, GRPO calculate the importance ratio $w_{i,t}(\theta)$ and the estimated advantage $\widehat{A}_{i,t}$ for every individual token $y_{i,t}$, specifically, the same advantage value $\widehat{A}_{i}$ is shared across all tokens within $y_{i}$. The GRPO loss is defined as:
\begin{equation}
\begin{aligned}
&\mathcal{J}_{\mathrm{GRPO}}(\theta)
=
\mathbb{E}_{x \sim \mathcal{D},\;
\{y_i\}_{i=1}^G \sim \pi_{\theta_{\text{old}}}(\cdot|x)}
\\
&\qquad\qquad
\bigg[
\frac{1}{G}
\sum_{i=1}^{G}
\frac{1}{|y_i|}
\sum_{t=1}^{|y_i|}
\min \Big(
w_{i,t}(\theta)\, \widehat{A}_{i,t},
\mathrm{clip}\big(w_{i,t}(\theta), 1-\varepsilon, 1+\varepsilon\big)\, \widehat{A}_{i,t}
\Big)
\bigg],
\end{aligned}
\end{equation}
where
\begin{equation}
    \begin{aligned}
w_{i, t}(\theta)=\frac{\pi_{\theta}\left(y_{i, t} \mid x, y_{i,<t}\right)}{\pi_{\theta_{\text {old }}}\left(y_{i, t} \mid x, y_{i,<t}\right)}, \quad \widehat{A}_{i, t}=\widehat{A}_{i}=\frac{r\left(x, y_{i}\right)-\text { mean }\left(\left\{r\left(x, y_{i}\right)\right\}_{i=1}^{G}\right)}{\operatorname{std}\left(\left\{r\left(x, y_{i}\right)\right\}_{i=1}^{G}\right)},
    \end{aligned}
\end{equation}

\textbf{Group Sequence Policy Optimization (GSPO).}
GSPO~\cite{zheng2025group} is a novel reinforcement learning algorithm designed for training large language models (LLMs), particularly Mixture-of-Experts (MoE) architectures. Unlike prior methods like GRPO that use token-level importance ratios, GSPO introduces sequence-level importance weighting. By defining the importance ratio based on the entire sequence likelihood and applying sequence-level clipping and optimization, GSPO significantly reduces gradient variance and training instability. This approach aligns the reward granularity with the optimization objective, leading to superior training efficiency, stability, and performance, as demonstrated in the latest Qwen3 models. The GSPO loss is defined as:
\begin{equation}
\begin{aligned}
&\mathcal{J}_{\mathrm{GSPO}}(\theta)
=
\mathbb{E}_{x \sim \mathcal{D},\;
\{y_i\}_{i=1}^{G} \sim \pi_{\theta_{\text{old}}}(\cdot|x)}
\\
&\qquad\qquad
\bigg[
\frac{1}{G}
\sum_{i=1}^{G}
\min \Big(
w_i(\theta)\, \widehat{A}_i,
\mathrm{clip}\big(w_i(\theta), 1-\varepsilon, 1+\varepsilon\big)\, \widehat{A}_i
\Big)
\bigg],
\end{aligned}
\end{equation}
where
\begin{equation}
    \begin{aligned}
      \widehat{A}_{i}=\frac{r\left(x, y_{i}\right)-\operatorname{mean}\left(\left\{r\left(x, y_{i}\right)\right\}_{i=1}^{G}\right)}{\operatorname{std}\left(\left\{r\left(x, y_{i}\right)\right\}_{i=1}^{G}\right)},
    \end{aligned}
\end{equation}
\begin{equation}
    \begin{aligned}
        w_{i}(\theta)=\left(\frac{\pi_{\theta}\left(y_{i} \mid x\right)}{\pi_{\theta_{\text {old }}}\left(y_{i} \mid x\right)}\right)^{\frac{1}{\left|y_{i}\right|}}=\exp \left(\frac{1}{\left|y_{i}\right|} \sum_{t=1}^{\left|y_{i}\right|} \log \frac{\pi_{\theta}\left(y_{i, t} \mid x, y_{i,<t}\right)}{\pi_{\theta_{\text {old }}}\left(y_{i, t} \mid x, y_{i,<t}\right)}\right).
    \end{aligned}
\end{equation}
The formula defines the GSPO (Generalized Sample-based Policy Optimization) objective $\mathcal{J}_{\text{GSPO}}(\theta)$ as the expectation over states $x$ drawn from the dataset $\mathcal{D}$ and sequences $y_i$ sampled from the old policy $\pi_{\theta_{\text{old}}}(\cdot|x)$, where for each sampled sequence $y_i$ of group size $G$, GSPO computes the average (over $G$ sequences) of the minimum between the policy ratio $w_i(\theta)$ and its clipped version $\text{clip}(w_i(\theta), 1-\varepsilon, 1+\varepsilon)$, both multiplied by the normalized advantage $\hat{A}_i$, itself computed as the difference between the reward $r(x, y_i)$ and the mean reward across all $G$ sequences, divided by their standard deviation; the importance ratio $w_i(\theta)$ is then given as the geometric mean (or equivalently, the exponential of the average log-ratio) of the per-timestep policy ratios, where $y_{i,<t}$ denotes the prefix of the sequence up to but not including timestep $t$, and $|y_i|$ is the length of the sequence.

\begin{figure*}[t]
\centerline{\includegraphics[width=0.95\columnwidth]{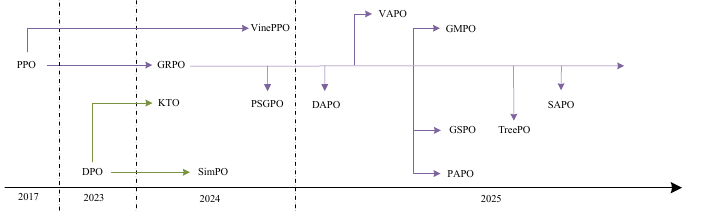}}
\caption{Evolution diagram of RL algorithm technology.}
\label{rader}
\end{figure*} 

\textbf{Decouple Clip and Dynamic sAmpling Policy Optimization (DAPO).}
DAPO~\cite{yu2025dapo}  is an advanced reinforcement learning algorithm for large language models (LLMs), proposed by ByteDance and Tsinghua University. Building upon GRPO, it introduces four key innovations: Clip-Higher (asymmetric clipping to boost exploration), Dynamic Sampling (filtering invalid samples to improve efficiency), Token-level Policy Gradient Loss (balancing token contributions), and Overlong Reward Shaping (penalizing excessive length). By removing the KL divergence constraint and the critic model, DAPO achieves superior performance in mathematical reasoning tasks, such as scoring 50 on AIME 2024 with only half the training steps compared to previous SOTA models. The DAPO loss is defined as:
\begin{equation}
\begin{aligned}
&\mathcal{J}_{\mathrm{DAPO}}(\theta)
=
\mathbb{E}_{x \sim \mathcal{D},\,\{y_i\}_{i=1}^{G} \sim \pi_{\theta_{\text{old}}}(\cdot \mid x)}
\\
&\qquad\qquad
\bigg[
\frac{1}{\sum_{i=1}^{G}|y_i|}
\sum_{i=1}^{G}
\sum_{t=1}^{|y_i|}
\min\big(
w_{i,t}(\theta)\,\hat{A}_{i,t},
\mathrm{clip}\big(w_{i,t}(\theta),\,1-\varepsilon_{\text{low}},\,1+\varepsilon_{\text{high}}\big)\,\hat{A}_{i,t}
\big)
\bigg]
\\
&\text{s.t.}\quad
0 < \big|\{y_i \mid \text{isEquivalent}(a, y_i)\}\big| < G,
\end{aligned}
\end{equation}
where
\begin{equation}
    \begin{aligned}
w_{i, t}(\theta)=\frac{\pi_{\theta}\left(y_{i, t} \mid x, y_{i,<t}\right)}{\pi_{\theta_{\text {old }}}\left(y_{i, t} \mid x, y_{i,<t}\right)}, \quad \hat{A}_{i, t}=\frac{R_{i}-\operatorname{mean}\left(\left\{R_{i}\right\}_{i=1}^{G}\right)}{\operatorname{std}\left(\left\{R_{i}\right\}_{i=1}^{G}\right)} .
    \end{aligned}
\end{equation}
Although reward models are often vulnerable to exploitation via reward hacking, a resilient substitute is adopted: leveraging the terminal accuracy of a verifiable task as the outcome reward, which is formally determined by the following criterion. Empirical efficacy in eliciting the deductive capabilities of base models is evident across diverse domains, such as software engineering and competitive mathematics; specifically, $p$ denotes the ground-truth solution and $\hat{p}$ signifies the prediction:
\begin{equation}
\begin{aligned}
R(\hat{p}, p) =
\begin{cases}
1,  & \mathrm{isEquivalent}(\hat{p}, p) \\
-1, & \text{otherwise}
\end{cases}
\end{aligned}
\end{equation}

\textbf{Soft Adaptive Policy Optimization (SAPO).}
SAPO~\cite{sapo} is an advanced reinforcement learning algorithm designed for large language models (LLMs), proposed by the Qwen team. It addresses the instability in policy optimization caused by high variance in token-level importance ratios, which is particularly severe in Mixture-of-Experts (MoE) models.
Unlike traditional methods like GRPO that rely on hard clipping, SAPO introduces a smooth, temperature-controlled gating function. This mechanism preserves useful learning signals while ensuring stability. Key innovations include a continuous trust region, sequence-level consistency, token-level adaptability, and an asymmetric temperature design that handles positive and negative advantages differently. This makes SAPO more robust and efficient for training complex reasoning models like Qwen3-VL. Adhering to the definitions in GRPO, SAPO defines both $\widehat{A}_{i, t}$ and $w_{i, t}(\theta)$; concurrently, $\tau_{\mathrm{pos}}$ and $\tau_{\mathrm{neg}}$ serve as the temperature hyperparameters governing positive and negative tokens, respectively, where $\sigma(x) = 1 /\left(1+e^{-x}\right)$ represents the sigmoid activation function.
The SAPO loss is defined as:

\begin{equation}
    \begin{aligned}
        \mathcal{J}_{\mathrm{SAPO}}(\theta)=\mathbb{E}_{q \sim \mathcal{D},\left\{y_{i}\right\}_{i=1}^{G} \sim \pi_{\theta_{\text {old }}}(\cdot \mid q)}\left[\frac{1}{G} \sum_{i=1}^{G} \frac{1}{\left|y_{i}\right|} \sum_{t=1}^{\left|y_{i}\right|} f_{i, t}\left(w_{i, t}(\theta)\right) \widehat{A}_{i, t}\right],
\end{aligned}
\end{equation}
where
\begin{equation}
\begin{aligned}
f_{i,t}(x)
&=
\sigma\!\left(\tau_{i,t}(x-1)\right) \cdot \frac{4}{\tau_{i,t}},
\\
\tau_{i,t}
&=
\begin{cases}
\tau_{\mathrm{pos}}, & \text{if } \widehat{A}_{i,t} > 0 \\
\tau_{\mathrm{neg}}, & \text{otherwise}.
\end{cases}
\end{aligned}
\end{equation}

\textbf{Other RLs.}
Beyond the widespread adoption of Direct Preference Optimization (DPO), an offline algorithm valued for simplifying Reinforcement Learning from Human Feedback (RLHF) and enhancing training stability, emerges SimPO~\cite{meng2024simpo}, a method distinguished by its reliance on the average log probability of a sequence as an implicit reward. This design, which eschews the need for a reference model, not only boosts computational efficiency but also exemplifies a broader family of objectives. Termed Human-Aware Losses (HALOs), this family includes approaches like Kahneman-Tversky Optimization (KTO)~\cite{ethayarajah2024model}, which leverages a behavioral model of human utility to directly maximize the utility of generations rather than preference log-likelihoods. 
Some evaluation reveals severe shortcomings in standard approaches like PPO, demonstrating that value networks often yield poor estimates of expected returns, barely surpassing random baselines on reasoning-heavy tasks. Addressing this deficiency, researchers introduce VinePPO~\cite{kazemnejad2024vineppo}, a straightforward yet potent algorithm that exploits the flexibility of language environments to derive unbiased Monte Carlo estimates. 
Achieving state-of-the-art performance on the AIME 2024 benchmark, VAPO~\cite{yue2025vapo} emerges as a novel framework specifically tailored for reasoning models, building upon the foundation of PPO. Concurrently, while Reinforcement Learning (RL) with unit test feedback has boosted LLM code generation, its reliance on sparse, post-hoc rewards restricts learning efficiency, particularly when all tests fail and no signal is received. Addressing this limitation, PSGPO~\cite{dai2024process} is proposed to shift the paradigm from outcome-based to process-based supervision. By delivering dense, line-level feedback that emulates human refinement during generation, PSGPO provides immediate guidance, thereby facilitating incremental improvements on complex coding tasks. 
The instability of GRPO during policy updates, often triggered by outlier tokens exhibiting extreme importance ratios, is effectively mitigated by GMPO~\cite{zhao2025geometric}. By shifting the optimization objective from the arithmetic to the geometric mean of token-level rewards, GMPO offers a plug-and-play stabilization mechanism that is inherently robust to reward outliers. Complementing this, TreePO~\cite{li2025treepo} revolutionizes the rollout phase by conceptualizing generation as a tree-structured search; its dynamic sampling policy utilizes local uncertainty to spawn additional branches, addressing the exploration limitations of costly on-policy rollouts. Further extending the RL paradigm beyond text, PAPO~\cite{wang2025perception} confronts the suboptimality of standard RLVR in multimodal contexts. It introduces a novel policy gradient algorithm that unifies perception and reasoning, enabling models to learn both "what to see" and "how to reason" simultaneously.

\section{Challenges}

\subsection{Environments}
Addressing a central bottleneck in agent development requires moving beyond the traditional conception of training environments as static fixtures. Instead, a nascent but crucial frontier for Agentic RL is to treat these environments as dynamic systems subject to active optimization. 
A novel reinforcement learning work, RLAnything~\cite{wang2026rlanything}, automatic environment adaptation, motivated by theoretical principles, is leveraged to refine both reward and policy models, enabling experiential learning driven by critic feedback from each component. Dynamically assess task difficulty through the accuracy of strategy deduction, guide language model task modification through error analysis of the reward model, and ensure effective modification through quality control.

\subsection{Trustworthy AI}
To mitigate this misinformation, a topology-guided security framework named G-Safeguard~\cite{wang2025gsafeguard} is introduced, employing Graph Neural Networks to detect anomalies on the multi-agent utterance graph and performing topological interventions for robust attack remediation.
Beyond multi-agent systems, Large Language Models are prone to hallucinations~\cite{cossio2025comprehensive,song2025hallucination,yao2025hallucination}.
Despite the significant advancement of Large Language Models (LLMs) in reasoning tasks via Reinforcement Learning (RL) optimization, a critical drawback emerges: the prevalence of hallucinations is markedly increased by reasoning-oriented RL fine-tuning. To counteract this, FSPO~\cite{li2025reasoning} is proposed. By leveraging automated verification against given evidence to dynamically adjust token-level advantage values, FSPO incentivizes factual correctness throughout the reasoning process, thereby mitigating the identified training pathologies.

\subsection{Capability Boundaries}
Despite the significant advancement of complex reasoning abilities via Reinforcement Learning with Verifiable Reward (RLVR)~\cite{yue2025does}, a fundamental barrier persists: the inherent capability boundaries of the base LLM remain largely unbreached due to on-policy constraints, vast action spaces, and sparse rewards. To overcome this impasse, RL-PLUS~\cite{dong2025rl} is proposed, a novel hybrid-policy optimization framework that synergizes internal exploitation with external data. By integrating Multiple Importance Sampling to rectify distributional mismatch and an Exploration-Based Advantage Function to navigate high-value reasoning paths, RL-PLUS empowers LLMs to transcend the limitations of their base models.
Addressing the distinct challenge of long-horizon, sparse-reward multi-turn decision-making, where flat policies suffer from unstable credit assignment, the HiPER framework~\cite{peng2026hiper} is proposed. By factorizing the agent into a high-level planner and a low-level executor, HiPER facilitates hierarchical optimization via a novel technique called Hierarchical Advantage Estimation (HAE). Complementing this, the fundamental bottleneck of context length explosion in tool-use scenarios is tackled by a summarization-based training regime~\cite{lu2025scaling}. This approach compresses tool usage histories into task-relevant summaries, enabling the derivation of a policy gradient representation that seamlessly optimizes both tool-use behaviors and summarization strategies within a compact, end-to-end RL infrastructure.
In work~\cite{yu2025demystifying}, key insight reveals that {real end-to-end tool-use trajectories} significantly outperform stitched synthetic data for SFT initialization, while high-diversity, model-aware datasets are shown to sustain exploration and markedly boost RL performance. Furthermore, the adoption of exploration-friendly techniques, specifically clip higher, overlong reward shaping, and entropy maintenance, is identified as crucial for training efficiency. Finally, a deliberative strategy characterized by fewer tool calls is demonstrated to surpass both verbose self-reasoning and frequent tool invocation, enhancing both tool efficiency and final accuracy. 
Advancing the paradigm of Agentic Reinforcement Learning, researchers introduce StepPO~\cite{wang2026steppo} as a definitive position on step-level optimization.

\subsection{System}
A simple yet profound observation underpins the OpenClaw-RL~\cite{wang2026openclaw}: the next-state signals generated by every agent interaction, be it a user reply, tool output, or GUI change, are universal, live learning sources largely ignored by existing systems. Rather than treating personal conversations, terminal executions, and SWE tasks as separate problems, a unified policy is trained within a single loop, simultaneously recovering two forms of information encoded in the next state. Evaluative signals, indicating action performance, are extracted as scalar rewards via a PRM judge, while directive signals, prescribing corrective action, are distilled through Hindsight-Guided On-Policy Distillation (OPD). By constructing an enhanced teacher context from textual hints, token-level directional advantage supervision richer than scalar rewards is provided. 
Addressing the trifecta of challenges rarely unified in GUI agent development, online training environments, rigorous benchmarking, and real-device deployment, the ClawGUI~\cite{clawgui} framework is introduced. Rather than treating these as disparate stages, ClawGUI provides an integrated lifecycle solution, establishing a cohesive pipeline that spans from reinforcement learning optimization to standardized evaluation and final production release.
Researchers introduce MetaClaw~\cite{metaclaw}, a continual meta-learning framework that unifies a base LLM policy with an evolving skill library through two mutually reinforcing mechanisms.

\begin{table}[h]
\centering
\caption{Comparison of previous works and this work. }
\label{tab:comparison}
\begin{tabular}{l c c c c}
\toprule
\textbf{Ref} & \textbf{LLMs} & \textbf{Agents} & \textbf{RL} & \textbf{Designs} \\
\midrule

~\cite{bai2025hallucination, fan2024survey, gao2024retrieval, han2025towards, li2025system, nguyen2025survey, szep2025fine, tao2024survey, tie2025survey, wang2024what, wu2025memory, zhang2024survey, singh2025agenticsurvey, liang2025reasoningsurvey, huang2025survey}& \checkmark &  &  & \\

~\cite{wu2025reinforcement, xiao2024comprehensive, li2025reinforcementsurvey, liu2025survey, pignatelli2024survey} &  &  &  \checkmark  & \\

~\cite{gao2025survey,masterman2024landscape,aratchige2025llms, dong2025survey, huang2024understanding, ke2025survey, luo2025large, plaat2025agentic, wei2025planning, zhang2025surveygui}& \checkmark & \checkmark &    & \\

~\cite{luo2025large,xi2025survey}& \checkmark & \checkmark &    &  \\

~\cite{srivastava2025technical}&  &  &  \checkmark  &  \\

~\cite{zhou2025reinforced, wang2025code, wang2025survey, wang2024survey, guo2025survey, srivastava2025technical} & \checkmark &  &  \checkmark   & \\

~\cite{lin2025comprehensivesurvey,zhang2025landscape} & \checkmark & \checkmark &  \checkmark  & \\

\midrule

~\textcolor{forcePurple}{{[Ours]}}  & \checkmark & \checkmark &  \checkmark  & \checkmark \\

\bottomrule
\end{tabular}
\end{table}

\section{Conclusion} 
This paper provides a deep looking for exploring the emerging methods of Agentic Reinforcement Learning in LLMs. 
The analysis highlights how foundation models serve as both a catalyst and a substrate for these advanced agentic systems. While significant progress has been made, critical challenges remain, particularly in ensuring robust generalization, managing computational overhead, and establishing reliable evaluation metrics for open-ended tasks.

\bibliographystyle{ACM-Reference-Format}
\bibliography{sample-base}

\appendix

\newpage

\end{document}